\documentclass[sigconf, nonacm]{acmart} 
\usepackage{subcaption}
\usepackage{balance}
\usepackage{algorithm}
\usepackage{algpseudocode}
\usepackage{amsmath}
\usepackage{graphicx}
\usepackage{pdfpages}
\usepackage{booktabs}
\usepackage{tabularx}
\usepackage{float}   
\usepackage[justification=centering]{caption}
\usepackage{soul}

\AtBeginDocument{
  \providecommand\BibTeX{{
    \normalfont B\kern-0.5em{\scshape i\kern-0.25em b}\kern-0.8em\TeX}}}

\newcommand{\mb}[1]{\textcolor{red}{\textbf{MB}:#1}}

\begin{document}

\title{Utilizing Graph Neural Networks for Effective Link Prediction in Microservice Architectures%
\footnote{This paper has been accepted for presentation at ICPE 2025 and will be published in the ACM Digital Library as part of the conference proceedings.}}
\thanks{This paper has been accepted for presentation at ICPE 2025 and will be published in the ACM Digital Library as part of the conference proceedings.}

\author{Ghazal Khodabandeh}
\orcid{0009-0001-4587-1876}
\affiliation{%
  \institution{Brock University}
  \department{Computer Science}
  \streetaddress{1812 Sir Isaac Brock Way}
  \city{St. Catharines}
  \state{Ontario}
  \postcode{L2S3A1}
  \country{Canada}}
\email{gkhodobandeh@brocku.ca}

\author{Alireza Ezaz}
\orcid{0009-0001-4156-2750}
\affiliation{%
  \institution{Brock University}
  \department{Computer Science}
  \streetaddress{1812 Sir Isaac Brock Way}
  \city{St. Catharines}
  \state{Ontario}
  \postcode{L2S3A1}
  \country{Canada}}
\email{sezaz@brocku.ca}

\author{Majid Babaei}
\orcid{0000-0002-1394-4030}
\affiliation{%
  \institution{McGill University}
  \department{SCS \& ECE}
  \city{Montreal}
  \state{Quebec}
  \country{Canada}}
\email{majid.babaei@mcgill.ca}

\author{Naser Ezzati-Jivan}
\orcid{0000-0003-1435-6297}
\affiliation{%
  \institution{Brock University}
  \department{Computer Science}
  \streetaddress{1812 Sir Isaac Brock Way}
  \city{St. Catharines}
  \state{Ontario}
  \postcode{L2S3A1}
  \country{Canada}}
\email{nezzatijivan@brocku.ca}

\begin{abstract}

Managing microservice architectures in distributed systems is complex and resource-intensive due to the high frequency and dynamic nature of inter-service interactions. Accurate prediction of these future interactions can enhance adaptive monitoring, enabling proactive maintenance and resolution of potential performance issues before they escalate. This study introduces a Graph Neural Network (GNN)-based approach, specifically using a Graph Attention Network (GAT), for link prediction in microservice Call Graphs. Unlike social networks, where interactions tend to occur sporadically and are often less frequent, microservice Call Graphs involve highly frequent and time-sensitive interactions that are essential to operational performance.

Our approach leverages temporal segmentation, advanced negative sampling, and GAT’s attention mechanisms to model these complex interactions accurately. Using real-world data, we evaluate our model across performance metrics such as AUC, Precision, Recall, and F1 Score, demonstrating its high accuracy and robustness in predicting microservice interactions. Our findings support the potential of GNNs for proactive monitoring in distributed systems, paving the way for applications in adaptive resource management and performance optimization.
\end{abstract}




\maketitle

\label{intro}
\section{Introduction}

In distributed systems composed of numerous interdependent microservices, ensuring reliability is crucial to maintaining uninterrupted operations \cite{lindenmayer2009adaptive}. The decentralized nature of microservice architectures, where smaller, independent services handle specific tasks, introduces unique challenges compared to monolithic systems. These challenges arise from the highly dynamic and interconnected dependencies between services, where disruptions in one component can cascade throughout the system. Traditional heuristic approaches often fall short in predicting and managing such complex interactions. Machine learning (ML), with its ability to analyze large-scale, dynamic data, offers a more effective solution by identifying critical patterns and enabling proactive maintenance \cite{li2021understanding, ruberto2022experimental}. As microservice interactions grow in complexity and scale, ML-based methods are essential for ensuring robust and adaptive system performance.


An effective way to represent these interactions is through Call Graphs, where nodes represent microservices and edges depict data exchanges between them \cite{findik2024using}. Although Call Graphs share some characteristics with social networks, they differ significantly due to their high-frequency, operationally driven, and temporally sensitive interactions—factors that are crucial for performance in distributed systems. This temporal aspect, along with features like response times and resource usage metrics, introduces unique challenges, requiring models that can handle dense, time-sensitive connections.

In this context, link prediction within Call Graphs provides a valuable way to anticipate future microservice interactions, allowing systems to proactively address potential performance issues. Traditional link prediction methods have proven effective in fields like social networks \cite{liben2007link, backstrom2011supervised} and recommendation systems \cite{sarwar2001item, lu2011link}. However, they often struggle with the densely connected, dynamic characteristics of microservices, where interactions shift continuously over time.

GNNs offer an advanced solution by simultaneously learning from the structural and temporal dynamics of Call Graphs \cite{he2023graphgru}. GNNs' dual capability makes them especially suitable for capturing complex, evolving relationships in microservices, where dependencies shift in response to changing operational demands. While GNNs have seen success in other domains, such as recommendation systems and biology, they are relatively underutilized in microservice monitoring, leaving an opportunity to apply GNN-based link prediction to improve system resilience \cite{ma2019graph}.

This study proposes a GNN-based framework for link prediction specifically tailored to microservice Call Graphs. Our contributions are as follows:  

\begin{enumerate}  
    \item We develop a temporal segmentation approach to address the challenge of capturing evolving interactions within highly dynamic microservice networks. This enables the model to learn both short-term and long-term dependencies, improving its ability to track interaction patterns and anticipate future interactions.  

    \item To tackle the issue of class imbalance, where negative samples far outnumber positive ones, we implement an advanced negative sampling strategy. This improves predictive accuracy by ensuring the model is trained on challenging and meaningful data points.  

    \item We perform an extensive evaluation using real-world microservice data to validate the practical applicability of our framework. This demonstrates the robustness of the model and its effectiveness in proactive monitoring, addressing the challenge of adapting to the dynamic and complex nature of distributed systems.  
\end{enumerate}

Our framework demonstrates GNNs' effectiveness in forecasting microservice interactions and offers a scalable approach that can be adapted for similar predictive tasks in other complex networks.

The paper is organized as follows: Section 2 reviews related work and foundational concepts. Section 3 outlines our methodology for adapting GNNs to link prediction in Call Graphs. Section 4 presents an experimental evaluation with real-world data, and Section 5 concludes with a summary of findings and future research directions for advancing adaptive monitoring in distributed systems.

\label{background}
\section{Related Works and Background}
The rise of complex software systems has driven a shift towards microservice architectures, where applications are decomposed into smaller, autonomous services \cite{velepucha2023survey}. Each service is designed to perform a specific function and communicates with others through well-defined APIs, creating a complex network of interactions. Understanding these interactions is critical for effective microservice management, enabling performance optimization and resource allocation \cite{findik2024using}.

This section reviews link prediction techniques in microservice call graphs, covering microservice architecture, call graph dynamics, and progression from traditional methods to GNNs, which effectively capture complex, evolving interactions. This background establishes the foundation for our GNN-based approach tailored to microservice link prediction.

\subsection{Link Prediction in Microservice Architectures}
Microservices architecture involves building applications from independent services that interact through APIs, supporting modularity and scalability \cite{velepucha2023survey}. Unlike monolithic systems, microservices are crucial in distributed environments, where components coordinate to maintain system functionality \cite{li2021understanding}. Link prediction in this context enhances load balancing, optimizes resource allocation, and prevents bottlenecks, ensuring high performance and resilience \cite{ruberto2022experimental}.


\subsection{Microservice Call Graphs (CallGraphs)}
Microservice architectures rely on CallGraphs to capture the interactions between services, providing a real-time view of dependencies. In these graphs, nodes represent individual microservices, while edges represent the communication or data exchanges between them. Each interaction is associated with a timestamp, which records the precise moment when the communication occurs, allowing the model to track the temporal dynamics of service dependencies. Unlike traditional directed acyclic graphs (DAGs), CallGraphs in microservices often follow a heavy-tailed distribution, with a few highly connected nodes and many sparsely connected ones \cite{tam2023pert}. These graphs are dynamic, as microservices continuously update their interactions, creating unique challenges for link prediction due to the need to account for real-time dependencies and the constantly evolving nature of service interactions.


\subsection{Overview of Link Prediction Techniques}
Link prediction involves forecasting potential future links by analyzing patterns and structures within a network graph \cite{hasan2011survey}. Traditional techniques for link prediction fall into three main categories: similarity-based methods, dimensionality reduction methods, and machine learning approaches \cite{arrar2024comprehensive}. Each offers distinct advantages:

\begin{itemize}
    \item \textbf{Similarity-Based Methods:} Techniques such as Random Walks and community detection estimate link probabilities by capturing node proximity and community structures \cite{liben2007link}.
    \item \textbf{Dimensionality Reduction Methods:} Embedding and matrix factorization methods simplify graph complexity, often using latent space models to reveal potential connections in dynamic social networks \cite{sarkar2005dynamic}.
    \item \textbf{Machine Learning Approaches:} These include supervised and unsupervised learning techniques, with advanced methods like GNNs that capture complex relationships and temporal dynamics within graph structures \cite{kipf2017semi}.
\end{itemize}

While traditional methods have proven effective in domains such as social networks and recommendation systems, the dynamic and complex nature of microservices demands a more robust approach. 

\subsection{GNNs for Link Prediction}
GNNs are well-suited to model the intricate, evolving relationships inherent in CallGraphs by combining structural and temporal data, making them particularly applicable to the complex, dynamic nature of microservice architectures \cite{tam2023pert, wang2023multi}. GNNs aggregate information from neighboring nodes and the broader graph structure, capturing both local and global dependencies essential for accurate link prediction \cite{xue2022dynamic}.

Though GNNs have seen widespread applications in fields like recommendation systems \cite{wang2021incorporating}, biology \cite{kang2022lr}, and social networks \cite{wu2022link}, their use in microservices is relatively underexplored. For microservices, GNNs can offer precise link predictions by modeling both individual microservice behaviors and the network-wide patterns of CallGraphs.

Our approach leverages GNNs due to their robustness in representing intricate, evolving relationships within CallGraphs \cite{tam2023pert}. We employ temporal GNN variants, such as Temporal Graph Attention Networks (TGAT) and Temporal Graph Networks (TGN), to capture the evolving nature of microservice interactions. Additionally, we integrate advanced negative sampling to improve predictive accuracy in sparse datasets, particularly by focusing on challenging negative samples. These adaptations make GNNs especially suitable for capturing nuanced, time-sensitive dependencies within microservice-based CallGraph networks, highlighting their relevance in this complex context \cite{wang2023multi}.

\begin{figure*}[htbp]
    \centering
    \includegraphics[width=\textwidth, trim={1cm 6.5cm 0cm 2.5cm}, clip]{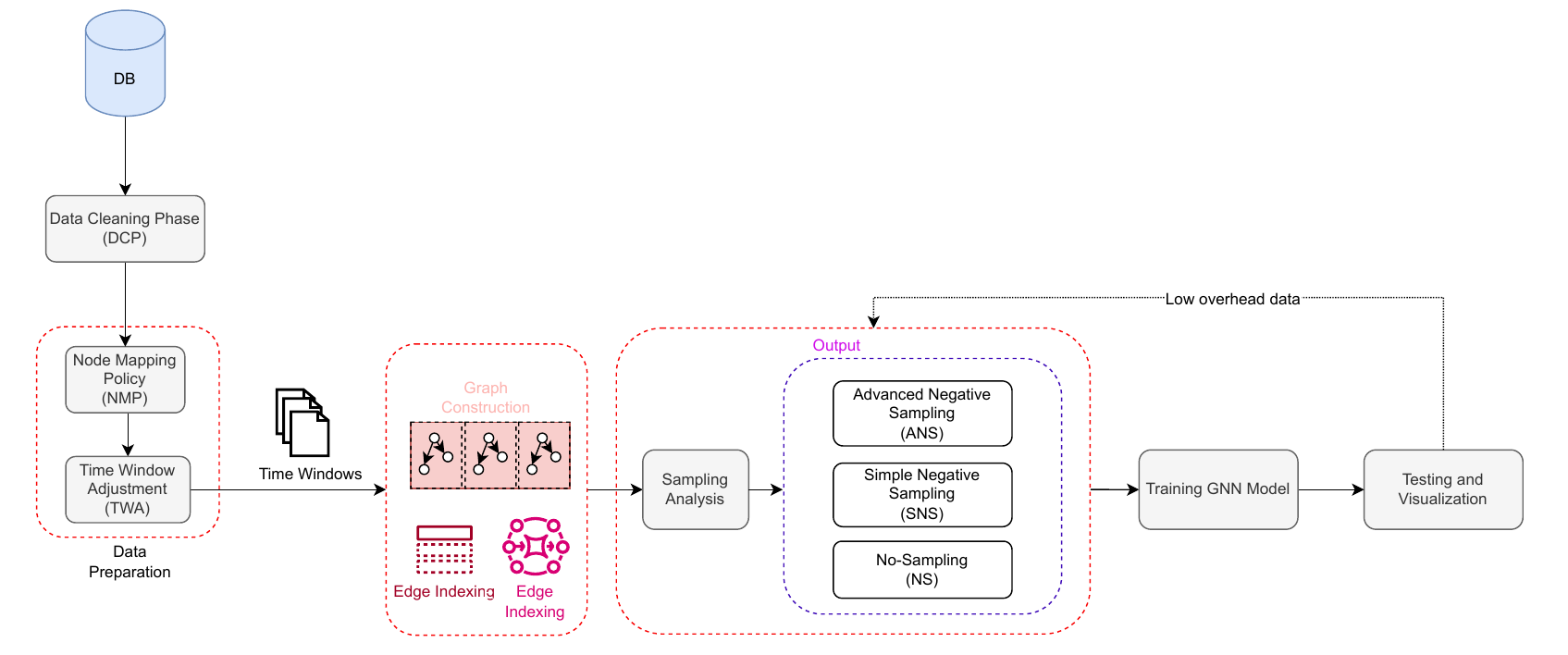}
    \caption{Methodology Diagram}
    \label{fig:diagram}
\end{figure*}

\label{methods}
\section{Methodology}

Our methodology uses link prediction to improve adaptive monitoring in microservice environments, enabling early detection of high-risk interactions and optimizing resource allocation \cite{lindenmayer2009adaptive}. By forecasting microservice connections, we can proactively address bottlenecks and enhance system performance. GNNs are ideal for this task due to their ability to model complex and dynamic network structures within microservice architectures.


As illustrated in Figure~\ref{fig:diagram}, our approach begins with data preparation, where we apply a Node Mapping Policy (NMP) to uniquely encode nodes, and a Time Window Adjustment (TWA) to segment data according to temporal intervals, capturing the evolution of microservice interactions. Each time window produces an independent graph, with interactions represented as edge indices and features. 

To address class imbalance in the dataset—a common challenge in link prediction— we perform a sampling analysis to determine the optimal sampling technique: advanced, simple, or none. Advanced negative sampling is selected to counterbalance the limited positive samples, improving the model’s precision in distinguishing true connections \cite{kotnis2017analysis}.

The core predictive model utilizes a GAT that learns from these temporal graphs by assigning attention weights to key connections. This mechanism refines node feature aggregation, helping the model prioritize influential interactions. During training, link probabilities are iteratively computed and optimized through binary cross-entropy loss, with backpropagation refining the model’s accuracy in distinguishing true links.

Testing then applies the trained model to unseen data, predicting potential future links. We use visualizations such as attention heatmaps, confusion matrixes, and precision-recall curves to analyze link patterns and capture insights on evolving microservice interactions over time.


\subsection{Data Description}

In this study, we use a trace of microservice events as our dataset. This trace data, denoted as \( T \), captures directed interactions between microservices over time. Each event in the trace is represented as a tuple containing a caller, a callee, a timestamp, and additional attributes, formally defined as follows:

\begin{equation}
T = \{ (s_i, d_i, t_i, A_i) \mid s_i, d_i \in S, \, t_i \in \mathbb{R}, \, A_i \in \mathbb{R}^k \}
\end{equation}

where:
\begin{itemize}
    \item \( S \) is the set of unique microservices in the system.
    \item Each tuple \( (s_i, d_i, t_i, A_i) \in T \) represents a single interaction:
    \begin{itemize}
        \item \( s_i \): Source microservice (caller) initiating the interaction.
        \item \( d_i \): Destination microservice (callee) receiving the interaction.
        \item \( t_i \): Timestamp of the interaction, recorded as a real number in Unix time format.
        \item \( A_i \): A vector of attributes associated with the interaction, including information such as service name, RPC type, trace ID, and response time.
    \end{itemize}
\end{itemize}
This dataset captures a directed sequence of interactions between microservices, represented by:
\begin{itemize}
    \item Direction: Each interaction specifies a caller (\( s_i \)) and a callee (\( d_i \)).
    \item Temporal Context: The timestamp (\( t_i \)) supports temporal analysis, allowing segmentation of interactions over specific intervals.
    \item Attributes: Each event also includes additional attributes (\( A_i \)), such as service name, latency, and RPC type, which are not analyzed in this study.
\end{itemize}

This trace data provides the foundation for constructing a graph representation of microservice interactions, where each microservice is represented as a node, and each interaction forms a directed edge. This graph structure allows us to analyze the temporal and structural evolution of microservice relationships, which will be explored in detail in subsequent sections.

\subsection{Preprocessing}

The collected raw trace data undergoes several preprocessing steps to transform it into a format suitable for temporal graph analysis. These steps include data cleaning, node mapping, and time window segmentation, each refining the dataset by filtering out irrelevant data, standardizing node identifiers, and organizing interactions into time-based segments. The preprocessing approach is summarized in Algorithm \ref{alg:Data Preparation}.

First, in the \textbf{Data Cleaning Phase (DCP)}, irrelevant or noisy trace data are removed to ensure that only valid and meaningful records remain. The cleaned dataset \( T_{clean} \) is defined as:

\begin{equation}
T_{clean} = \{ (s_i, d_i, t_i, A_i) \in T \mid \text{isRelevant}(s_i, d_i, t_i, A_i) = \text{true} \}\end{equation}

where \textbf{isRelevant} is a filtering function that removes incomplete records or interactions outside a valid time range. After this cleaning process, the traces in \( T_{clean} \) are sorted by timestamp to prepare the data for temporal analysis.

Next, a \textbf{Node Mapping Policy (NMP)} is applied to each unique microservice in \( T_{clean} \), converting raw identifiers into standardized node IDs. This produces a mapped dataset \( T_{map} \):

\begin{equation}
T_{map} = \{ (f(s_i), f(d_i), t_i, A_i) \mid (s_i, d_i, t_i, A_i) \in T_{clean} \}
\end{equation}

Here, \( f(s_i) \) and \( f(d_i) \) represent the mapped identifiers for each caller and callee microservice, respectively. This transformation assigns each unique microservice a consistent node ID, which simplifies both graph construction and analysis. Additionally, unique nodes from both callers and callees are collected to avoid conflicts in the mapping process, ensuring consistency across the graph.

To capture temporal dynamics, we segment the cleaned and mapped interactions into fixed Time Windows. Using fixed windows instead of sliding windows provides a clear structure for segmenting interactions, minimizing computational complexity and avoiding data redundancy. Each time window represents a defined interval, allowing the model to analyze changes over time and gain a clear view of how microservice interactions evolve across distinct periods. For example, a microservice handling user requests may show high interaction frequency with a database service during peak usage times and much lower activity during off-peak hours. This fixed-window approach enables the model to learn sequential, time-dependent relationships, which is essential in microservice architectures where interaction patterns vary over time. The time-windowed dataset \( T_w \) is defined as:

\begin{equation}
T_w = \{ T_{w_j} \mid T_{w_j} = \{ (f(s_i), f(d_i), t_i, A_i) \in T_{map} \mid t_i \in w_j \} \}
\end{equation}

where:
\begin{itemize}
    \item \( W = \{w_1, w_2, \ldots, w_n\} \) represents the set of fixed time intervals (e.g., 0–100 ms, 100–200 ms).
    \item Each \( T_{w_j} \) subset contains interactions within a specific time window \( w_j \).
\end{itemize}

After these preprocessing steps, the dataset \( T_w \) is ready for conversion into a graph representation, which enables temporal graph analysis of microservice interactions, as shown in Figure~\ref{fig:diagram}.
\begin{algorithm}[H]
    \caption{Data Preparation}
    \label{alg:Data Preparation}
    \begin{algorithmic}[1]

        \State \textbf{Get Unique Nodes:} 
        \State $nodes \gets \text{UNIQUE(CONCAT}(df['um'], df['dm']))$
        
        \State \textbf{Map and Encode Nodes:} 
        \State $map \gets \{n: i \, | \, i, n \in \text{ENUM}(nodes)\}$
        \State $df['um\_enc'] \gets \text{MAP}(df['um'], map)$
        \State $df['dm\_enc'] \gets \text{MAP}(df['dm'], map)$
        
        \State \textbf{Create Time Windows:} 
        \State $win \gets [\text{FILTER}(df, t, t + W_{size}) \, | \, t \in \text{RANGE}(0, T_{max}, W_{size})]$
        
        \State \textbf{Separate Train/Test:} 
        \State $train \gets \text{FILTER}(win, max\_t < T_{train})$
        \State $test \gets \text{FILTER}(win, T_{train} \leq min\_t < T_{max})$
    \end{algorithmic}
\end{algorithm}
\subsection{Graph Construction}

In our method, we construct a directed graph for each time window to capture interactions between microservices as they evolve over time. This time-windowed graph construction allows us to model dynamic interactions in a structured and analyzable format \cite{iyer2016time}. By representing each microservice as a node and each interaction as a directed edge, we capture the changing communication patterns within the system. This approach preserves the temporal context of interactions, enabling advanced analysis techniques, such as link prediction, using GNNs.

By creating individual graphs for distinct temporal segments, as illustrated in Algorithm~\ref{alg:Graph Construction}, we can analyze how relationships and behavior within the microservice system change across periods, maintaining the temporal integrity of the data.

Formally, let \( V = \{ v \mid v = f(s), \, s \in S \} \) represent the set of nodes, where each node \( v \in V \) corresponds to a unique microservice identifier derived from the node mapping function \( f \) described in preprocessing. The set of directed edges \( E_{w_j} \) for each time window \( w_j \) is defined as:
$E_{w_j} = \{ (v_s, v_d) \mid (f(s_i), f(d_i), t_i) \in T_{w_j} \}$

where each edge \( (v_s, v_d) \) indicates an interaction from source node \( v_s \) (caller) to destination node \( v_d \) (callee), with \( t_i \) lying within the specified time window \( w_j \).

To represent the direction of each interaction, we encode the edges in each graph \( G_{w_j} \) as pairs of indices, specifying which node is the caller (source) and which is the callee (destination). This encoding allows us to efficiently handle multiple interactions between the same nodes, helping the model to learn from repeated connections and capture complex patterns in microservice communications.

In keeping with the data simplification strategy introduced in preprocessing, we use an identity matrix as the sole feature representation for nodes. This matrix provides a straightforward indication of node identities without adding unnecessary complexity. Each diagonal element in the identity matrix is one, ensuring that each node has a unique, distinguishable representation. For our study, additional node or edge attributes are not included, as this basic representation effectively captures the necessary structural relationships.

\begin{algorithm}[H]
    \caption{Graph Construction}
    \label{alg:Graph Construction}
    \begin{algorithmic}[1]
        \Function{create\_graph}{df}
            \State $e\_index \gets \text{stack}(df['um\_enc'], df['dm\_enc'])$
            \State $e\_attr \gets \text{expand}(\text{tensor}(df['timestamp']))$
            \State \Return $\text{Data}(e\_index, e\_attr)$
        \EndFunction

        \For{$g \text{ in } train\_graphs + test\_graphs$}
            \State $g.x \gets \text{identity\_matrix}(n\_nodes)$
        \EndFor
    \end{algorithmic}
\end{algorithm}

\subsection{Sampling} 
Our approach addresses potential data imbalance within the dataset, where the number of positive connections (existing links) is significantly lower than the number of potential negative connections (non-links). In typical microservice interactions, the prevalence of non-links (0 labels) often exceeds that of existing connections (1 labels). However, in the specific context of network datasets like the CallGraph we are studying, only the existing connections are reported in the dataset. This results in an implicit imbalance, as unobserved links are treated as negative samples.

In this setting, including all possible negative samples would result in a dataset overwhelmingly skewed towards non-links, which would hinder the model’s ability to learn effectively. To address this, our sampling strategy carefully selects a balanced representation of negative samples, ensuring that the model can learn accurately from both observed and potential connections, thus enhancing its link prediction capabilities \cite{wang2024efficient}.

We evaluate three sampling strategies—No Sampling, Simple Negative Sampling, and Advanced Negative Sampling—each chosen based on the unique characteristics of the dataset. Algorithm \ref{alg:Advanced Negative Sampling} further details these methods.

\begin{enumerate}
    \item \textbf{No Sampling:} This approach is used when the dataset is balanced (i.e., \( |P| \approx |N| \), where \( P \) is the set of positive samples and \( N \) the set of negative samples). In such cases, the model is trained directly on the balanced dataset, preserving the original distribution.

    \item \textbf{Simple Negative Sampling:} For moderately imbalanced datasets (where \( |P| \ll |N| \)), we apply Simple Negative Sampling. This method selects a subset \( N_s \subset N \) of non-existent connections, ensuring \( |N_s| \approx |P| \) for a balanced dataset:
    \begin{equation}
    N_s = \{ (v_i, v_j) \mid v_i, v_j \in V, (v_i, v_j) \notin E \}
   \end{equation}
    This approach maintains computational efficiency by focusing only on a manageable number of negative samples.

    \item \textbf{Advanced Negative Sampling:} For datasets with significant imbalance or complex structures, Advanced Negative Sampling is applied. Here, the probability of selecting negative samples is weighted by node degree, capturing central interactions more accurately. For a node \( v \) with degree \( d_v \), the sampling probability is:
    \begin{equation}
    p(v) = \frac{d_v^{\alpha}}{\sum_{u \in V} d_u^{\alpha}}
    \end{equation}
    where \( \alpha \) is a tunable parameter controlling the influence of node centrality. Higher \( \alpha \) values prioritize nodes with more connections, better representing central hubs in the network.
\end{enumerate}

Our analysis indicated significant imbalance in the CallGraph dataset, leading us to select Advanced Negative Sampling with a tuned \( \alpha \) parameter. This approach explicitly excludes existing edges, ensuring that sampled pairs represent genuinely non-existing links. In dynamic call graphs, this additional check maintains the integrity of negative samples, which is essential for accurate training. Without this step, standard negative sampling might include pairs that are already connected, reducing the effectiveness of the model in identifying true potential connections. Algorithm \ref{alg:Advanced Negative Sampling} provides further implementation details for our enhanced sampling approach.

\begin{algorithm}
\caption{Advanced Negative Sampling}
\label{alg:Advanced Negative Sampling}
\begin{algorithmic}[1]
\Function{AdvNegSample}{e\_index, n\_nodes, ex\_edges, $\alpha=0.1$}
    \State degrees $\gets \text{count degrees from } e\_index$
    \State $deg\_prob \gets \text{normalize}((degrees / \text{sum}(degrees))^\alpha)$

    \State $ex\_set \gets \text{set of tuples from } ex\_edges$
    \State $neg\_edges \gets []$

    \For{each edge in $e\_index$}
        \Repeat
            \State $src \gets \text{sample node based on } deg\_prob$
            \State $dest \gets \text{rand from } 0 \text{ to } n\_nodes$
            \If{$(src, dest) \notin ex\_set$}
                \State \text{Add} $[src, dest] \text{ to } neg\_edges$
                \State \textbf{break}
            \EndIf
        \Until{valid neg edge found}
    \EndFor

    \State \Return \text{tensor}(neg\_edges)
\EndFunction
\end{algorithmic}
\end{algorithm}

\subsection{Model Characteristics}

In this study, we use a GAT as the core of our GNN to model the evolving relationships within the microservices call graph, focusing on predicting future links based on temporal interactions \cite{gu2019link}. GAT is particularly well-suited for this task due to its attention mechanism, which allows the model to selectively focus on critical connections in the network. Unlike Graph Convolutional Networks (GCNs) that assign equal importance to all neighboring nodes, GAT assigns variable importance to each connection, which is essential in microservice networks where certain interactions may carry more weight than others.

Our approach uses time-windowed graphs to capture the dynamic patterns across the network. Combining time-windowed segmentation with GAT allows the model to prioritize significant connections within each timeframe, supporting both temporal and structural analysis. Time-windowed graphs help track how microservice interactions change over time, while GAT’s attention mechanism highlights connections crucial for future link prediction.

Microservice architectures bring unique challenges in link prediction, characterized by high interaction frequency and dynamic changes in dependencies, which make them more complex than traditional social networks. The temporal aspect requires a model that can handle evolving patterns and heterogeneity in interactions. GAT’s multi-head attention mechanism addresses this by learning from multiple aspects of the input data simultaneously, enabling the model to adapt to the changing relationships typical in microservices.

During training, we compute link probabilities to differentiate between actual and potential connections over time. For each possible link, the model assigns a probability indicating the likelihood of the connection forming in the future. If this probability exceeds a set threshold, the link is classified as a likely future connection; otherwise, it is not. This probabilistic approach enables nuanced predictions about link formation. We use binary cross-entropy loss to optimize the model, improving accuracy in distinguishing actual connections from potential ones.

The GAT attention mechanism is formalized as follows:

\begin{equation}
\alpha_{ij} = \frac{\exp(\text{LeakyReLU}(a^T [W h_i || W h_j]))}{\sum_{k \in \mathcal{N}(i)} \exp(\text{LeakyReLU}(a^T [W h_i || W h_k]))}
\end{equation}

where \(\alpha_{ij}\) is the attention coefficient between nodes \(i\) and \(j\), \(W\) is the weight matrix applied to node features, and \(\mathcal{N}(i)\) is the neighborhood of node \(i\). This equation shows how the model prioritizes connections based on feature similarity.

For feature aggregation across neighbors in each attention head, we use:

\begin{equation}
h_i' = \sigma \left( \sum_{j \in \mathcal{N}(i)} \alpha_{ij} W h_j \right)
\end{equation}

where \( h_i' \) is the updated feature of node \( i \), and \( \sigma \) is an activation function, such as ReLU or ELU. This aggregation captures essential structural patterns within each neighborhood.

The multi-head attention mechanism further stabilizes the model by combining outputs from multiple heads:

\begin{equation}
h_i' = \parallel_{k=1}^K \sigma \left( \sum_{j \in \mathcal{N}(i)} \alpha_{ij}^{(k)} W^{(k)} h_j \right)
\end{equation}

where \( h_i' \) is the combined representation, \( \alpha_{ij}^{(k)} \) is the attention coefficient from the \( k \)-th head, and \( W^{(k)} \) is the weight matrix for each head. While computationally intensive, this approach is essential for nuanced predictions in dynamic microservice networks. 

\subsubsection{Model Construction}
Our GAT model includes adaptations specific to dynamic microservice interactions, with particular attention to temporal dependencies. It processes input features as a node feature matrix \( X \) and an edge index matrix \( E \), where \( X \) contains identity matrix values for nodes, and \( E \) specifies unique edges. Temporal information is embedded in both the node and edge features by incorporating timestamps within the node features, allowing the model to account for time-dependent interactions. The number of unique nodes determines the input feature space, denoted as \( input\_channels \).

The model comprises two graph attention layers. The first layer applies multi-head attention with two heads, capturing multiple perspectives of node interactions across time windows, followed by an Exponential Linear Unit (ELU) activation to learn complex patterns. The second layer consolidates features using a single head to focus on the most relevant outputs, reducing overfitting risk and improving computational efficiency. Temporal dependencies are implicitly captured through the inclusion of timestamp features, which affect the attention mechanism by adjusting the importance of edges over time.

Output embeddings represent relationships between microservices. The output dimension is selected to balance representational capacity and computational demands: higher dimensions capture nuanced relationships but increase complexity, while lower dimensions simplify the model at the cost of detail. We determined an optimal trade-off through experimentation.

The attention mechanism in our GAT model assigns variable importance to edges based on node features, which is critical in microservice environments where interaction importance varies over time. By tracking attention weights, we gain insights into key microservice interactions, with time-related aspects influencing edge importance.





\subsubsection{Training Loop}
Our training loop builds on the standard GNN training process, tailored to dynamic link prediction in microservice interactions. Initially, the model is set to training mode, and the optimizer’s gradients are reset at each iteration, ensuring that each weight update accurately reflects the current batch of data. This approach helps the model learn from evolving microservice dependencies and adapt to patterns unique to the dynamic nature of distributed systems.

During each iteration, the model processes input features to generate embeddings representing the microservices. By referencing the edge index, we identify source and destination nodes and calculate predicted probabilities for existing connections by applying a dot product followed by a sigmoid activation function. This provides the likelihood of each link, facilitating the model's ability to distinguish between actual and potential links.

To generate negative edge predictions, we use the same dot product and sigmoid method. After testing multiple loss functions, we determined that binary cross-entropy loss delivered optimal performance, which was essential in distinguishing between existing and non-existing connections accurately. Binary cross-entropy effectively handles the class imbalance in our dataset by minimizing errors for both positive and negative samples, a crucial factor for microservice networks with sparse observed connections and abundant potential ones. The binary cross-entropy loss function is defined as \cite{ruby2020binary}:

\begin{equation}
\mathcal{L} = - \frac{1}{|P| + |N|} \left( \sum_{(u,v) \in P} \log(\hat{y}_{uv}) + \sum_{(u,v) \in N} \log(1 - \hat{y}_{uv}) \right)
\end{equation}

where:
\begin{itemize}
    \item \( P \) is the set of positive edges,
    \item \( N \) is the set of negative edges,
    \item \( \hat{y}_{uv} \) represents the model’s predicted probability for each link.
\end{itemize}

The loss function evaluates the model’s performance on both observed and non-observed edges, reducing false positives while maintaining true positive rates, which is critical for dependable microservice interaction predictions.

Link prediction itself involves calculating a similarity score \( s_{ij} \) for each pair of nodes \( i \) and \( j \) as follows:

\begin{equation}
s_{ij} = h_i^T \cdot h_j
\end{equation}

where \( h_i \) and \( h_j \) are the embeddings of nodes \( i \) and \( j \), respectively. This similarity score \( s_{ij} \) is passed through a sigmoid function to yield a probability score:

\begin{equation}
p_{ij} = \sigma(s_{ij}) = \frac{1}{1 + e^{-s_{ij}}}
\end{equation}

This probability score \( p_{ij} \) represents the likelihood of an edge existing between nodes \( i \) and \( j \). A threshold \( \tau \) is applied for link classification:

\begin{equation}
p_{ij} > \tau \Rightarrow \text{Predict a link exists between } i \text{ and } j
\end{equation}

We chose a threshold \( \tau \) to balance precision and recall, as this ensures a good trade-off between identifying actual connections and minimizing false positives. This threshold-based classification enables a probabilistic yet binary decision regarding link existence in the call graph, enhancing the model’s robustness across dynamic time windows.

After calculating the loss, backpropagation computes gradients, and the optimizer updates the model parameters, refining the model's predictions over successive iterations. This iterative training process, coupled with binary cross-entropy and carefully tuned thresholding, ensures that our GNN model effectively captures and predicts evolving microservice interactions within the call graph.

\subsection{Testing and Visualization}
The testing procedure in our model involves applying the trained GNN to test windows in the CallGraph dataset, each representing a discrete time segment. This approach allows us to assess the model's link prediction capabilities across sequential time intervals. In each test window, we predict potential connections between nodes based on the learned embeddings and advanced negative sampling to generate contrasting (non-connected) pairs. The model outputs these predictions in the form of probability scores for each link, which we store for subsequent visual analysis.

For visualization, we generate Precision-Recall and ROC curves for each test window, saving them as images for easy comparison across time segments. Attention heatmaps illustrate the model's focus on specific node pairs, and a confusion matrix for the final test window provides an overview of prediction distribution. These visualizations facilitate an intuitive understanding of model behavior across testing intervals. We will discuss them more in the next section.

\label{results}
\section{Experimental Evaluation}
In this section, we evaluate the performance of our proposed approach on real-world microservice interaction data. We present the experimental setup, including the dataset and computational resources, followed by a detailed analysis of the model's effectiveness across various performance metrics. The results provide insights into our model's capability to handle dynamic microservice networks and its applicability to similar distributed system scenarios. \noindent {All related code, data, and results for this study are available at this link.\footnote{\url{https://github.com/ghazalkhb/Link_Prediction}}}

\subsection{Dataset and Experimental Setup}

\subsubsection{Dataset}
Our experiments use Alibaba’s 2022 Cluster Trace dataset for microservices \cite{luo2022depth}, which provides detailed runtime traces of approximately twenty thousand microservices over a thirteen-day period. This dataset captures real-world operational dynamics, making it suitable for modeling and predicting evolving microservice interactions.

\subsubsection{Training and Testing Split}
To ensure that the GNN model learns from historical interactions and accurately predicts future connections, we divided the data into distinct training and testing intervals. Specifically, we used timestamps from 0 to 7000 ms for training and 7000 to 10000 ms for testing, preventing any data leakage across these intervals. The number of records in the training phase is 589,540, while the testing phase includes 250,117 records. This setup allows the model to focus on learning temporal dependencies without bias from test data.

\subsubsection{System Specifications}
Our experiments ran on a system equipped with 51 GB of RAM and a 225.8 GB disk, with an average CPU load of 3.4 and approximately 62.7\% idle time, supporting efficient processing of the high-dimensional data. This configuration ensured that model training and evaluation could be conducted at scale, reflecting the practical requirements for deploying GNN-based predictions in real-world scenarios.

\subsection{Evaluation Metrics and Methodology}

To evaluate our approach, we use well-established metrics that cover both predictive performance and interpretability, as outlined in Table \ref{tab:evaluation_metrics} \cite{lu2011link}. 

\textbf{Performance Metrics} measure the model's prediction quality across various dimensions, including AUC, Accuracy, Precision, Recall, and F1 Score. These metrics collectively reveal how well the model discriminates between positive and negative links, maintains sensitivity to true connections, and balances predictive accuracy.

\textbf{Interpretability Metrics} provide insights into model behavior through the Precision-Recall Curve, ROC Curve, Confusion Matrix, and Attention Heatmap. These visual tools reveal the model's decision patterns, areas of focus, and prediction distribution, adding interpretative value to the quantitative metrics.

\begin{table}[h!]
    \centering
    \caption{Evaluation Metrics Summary}
    \begin{tabular}{|p{0.25\columnwidth}|p{0.65\columnwidth}|}
        \hline
        \textbf{Metric} & \textbf{Description} \\
        \hline
        \multicolumn{2}{|c|}{\textbf{Performance Metrics}} \\
        \hline
        AUC & Ability to separate positive and negative links. \\
        Accuracy & Proportion of overall correct predictions. \\
        Precision & Proportion of true positives among all positive predictions. \\
        Recall & Sensitivity to actual positive links. \\
        F1 Score & Harmonic mean of Precision and Recall. \\
        \hline
        \multicolumn{2}{|c|}{\textbf{Interpretability Metrics}} \\
        \hline
        Precision-Recall Curve & Shows trade-off between Precision and Recall. \\
        ROC Curve & Plots True Positive vs. False Positive Rate. \\
        Confusion Matrix & Breakdown of true positives, negatives, and errors. \\
        Attention Heatmap & Highlights model’s focus on connections. \\
        \hline
    \end{tabular}
    \label{tab:evaluation_metrics}
\end{table}

\subsection{Results and Analysis}

To evaluate performance, we compared our approach with a range of link prediction methods, each representing distinct perspectives on structural and temporal dynamics in network datasets. This comparison provides insights into how different methods address the complexities of evolving CallGraph data in microservice environments. Table~\ref{table:1} summarizes the comparative results, offering a quantitative overview of each method's performance across metrics such as Accuracy, Precision, Recall, F1 Score, and AUC, enabling a direct comparison of predictive capabilities within our CallGraph dataset.

\begin{table}[h]
    \centering
    \small 
    \caption{Comparison of Link Prediction Methods}
    \begin{tabular}{|l|c|c|c|c|c|}
        \hline
        \textbf{Method} & \textbf{AUC} & \textbf{Acc.} & \textbf{Prec.} & \textbf{Rec.} & \textbf{F1} \\ 
        \hline
        NodeSim & 0.50 & 0.60 & 0.40 & 0.17 & 0.18 \\
        \hline
        Adj. NodeSim & 0.62 & 0.69 & 0.39 & 0.20 & 0.10 \\ 
        \hline
        LSTM & 0.76 & 0.73 & 0.51 & 0.54 & 0.60 \\ 
        \hline
        Simple GNN & 0.94 & 0.69 & 0.62 & 0.97 & 0.76 \\ 
        \hline
        Simple Temp. GNN & 0.93 & 0.72 & 0.65 & 0.96 & 0.77 \\ 
        \hline
        Our Approach & 0.89 & 0.91 & 0.89 & 0.96 & 0.92 \\ 
        \hline
    \end{tabular}
    \label{table:1}
\end{table}

\subsubsection{Comparative Quantitative Results}
Table~\ref{table:1} provides a quantitative comparison across methods, allowing us to evaluate each model’s predictive capabilities on our CallGraph dataset. Among structure-driven approaches, NodeSim \cite{saxena2022nodesim} relies on structural similarity to estimate link probability, using random walks and node embeddings to capture network structure. While effective for static networks, NodeSim falls short in adapting to the dynamic interactions typical in microservice networks, achieving an AUC of 0.50 and an F1 score of 0.18. This underscores its limitations in capturing temporal changes. The adjusted variant of NodeSim, tailored to our dataset, improves in both AUC (0.62) and accuracy (0.69), though it still struggles with lower recall and precision. These results highlight the need for models that can incorporate temporal adaptability for dynamic microservice predictions.

For a temporal perspective, we benchmark our approach against a simple LSTM model, which is designed to capture sequential dependencies and time-based interactions within each window. The LSTM model achieves moderate performance, with an AUC of 0.76 and an F1 score of 0.60, and demonstrates a noticeable improvement in recall (0.54) over NodeSim. However, its lack of structural insight into node-specific relationships limits its predictive depth, reinforcing the advantage of combining both temporal and structural information for link prediction \cite{lu2011link}.

Both the Simple GNN and Simple Temporal GNN provide foundational insights into structural and temporal aspects of the data, respectively. While the Simple GNN captures static structural properties, the Simple Temporal GNN, with its basic time segmentation, performs better with an AUC of 0.93 and an F1 score of 0.77. These results indicate that incorporating temporal segmentation improves predictive performance, though both models lack the nuanced adaptability of our approach.

\begin{figure}
    \centering
    \begin{subfigure}{0.23\textwidth}
        \centering
        \includegraphics[width=\textwidth]{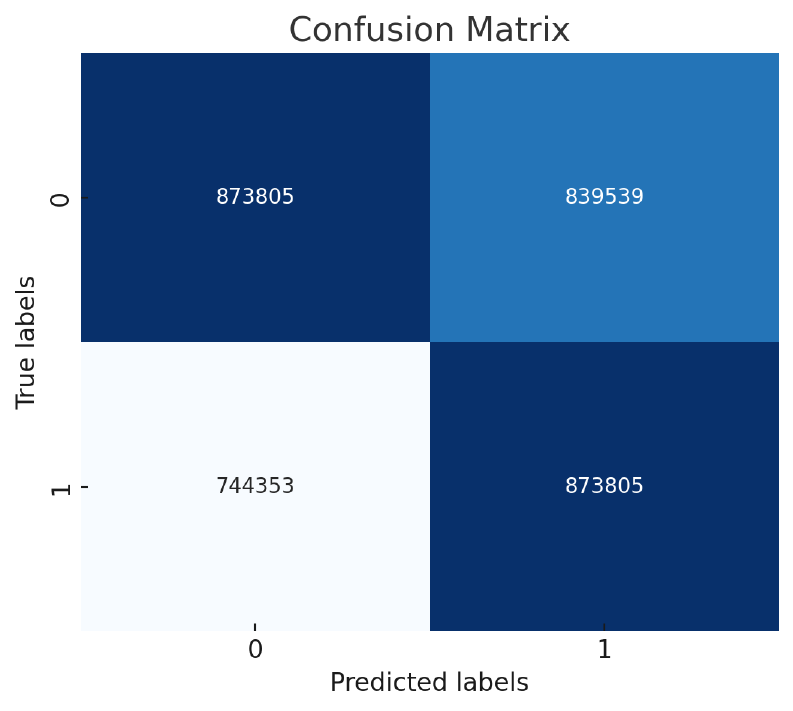}
        \caption{Confusion matrix for LSTM Approach}
        \label{fig:LSTM_C}
    \end{subfigure}
    \hfill
    \begin{subfigure}{0.23\textwidth}
        \centering
        \includegraphics[width=\textwidth]{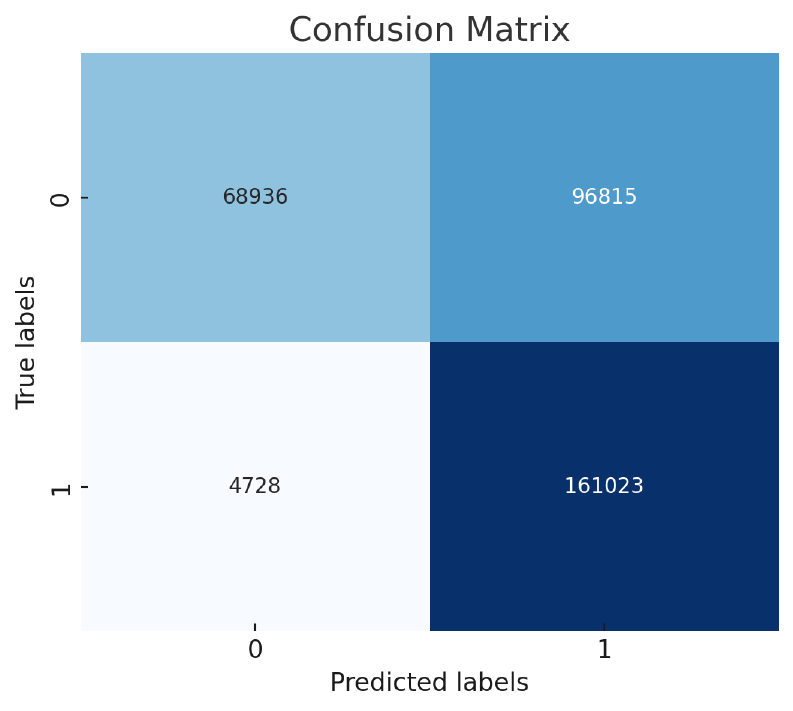}
        \caption{Confusion matrix for Simple GNN Approach}
        \label{fig:Simple_GNN_C}
    \end{subfigure}
    \vspace{1em}
    \begin{subfigure}{0.23\textwidth}
        \centering
        \includegraphics[width=\textwidth]{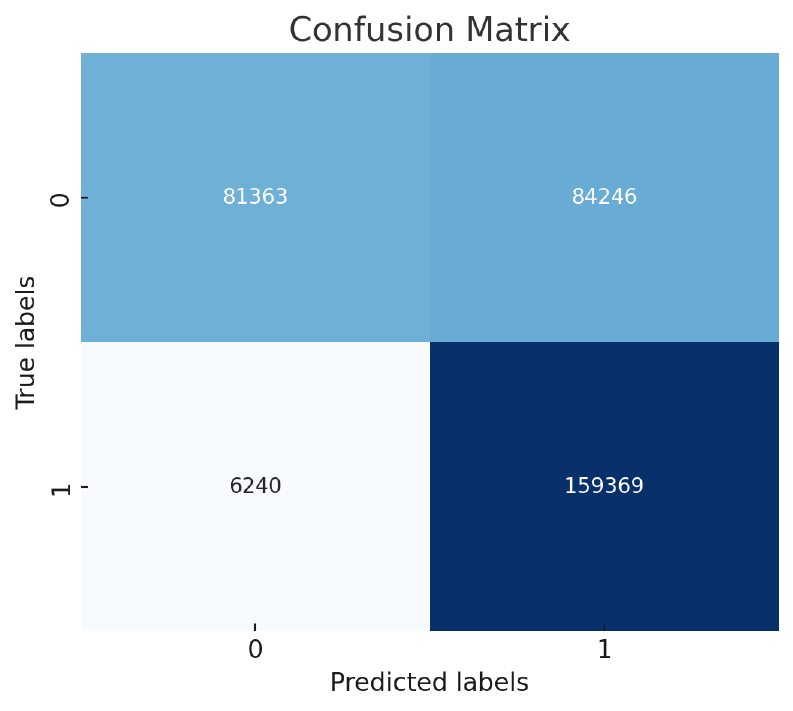}
        \caption{Confusion matrix for Simple Temporal GNN Approach}
        \label{fig:Simple_Temporal_GNN_C}
    \end{subfigure}
    \hfill
    \begin{subfigure}{0.23\textwidth}
        \centering
        \includegraphics[width=\textwidth]{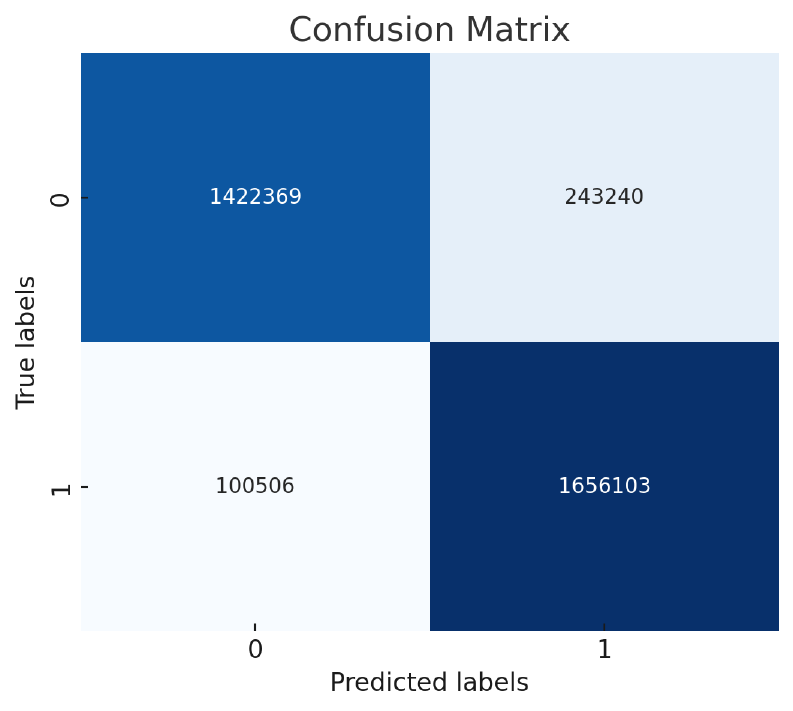}
        \caption{Confusion matrix for Our Approach}
        \label{fig:My_Approach}
    \end{subfigure}
    \caption{Confusion matrix illustrating the performance of different models: LSTM, Simple GNN, Simple Temporal GNN, and Our Approach. Each subfigure highlights the classification performance with their respective matrix.}
    \label{fig:confusion_matrices}
\end{figure}

\subsubsection{Model-Specific Analysis}
Our GAT-based approach, incorporating advanced negative sampling and temporal windowing, demonstrates strong predictive performance in microservice networks. Achieving an accuracy of 0.91 and an F1 score of 0.92, this model balances precision (0.89) and recall (0.96), effectively capturing the temporal dynamics and complex interactions within our dataset. Attention heatmaps generated during training reveal significant network connections, highlighting the model's ability to prioritize meaningful links while downplaying less relevant ones. Insights from the confusion matrix confirm this balance, showing a distribution of true positives and true negatives with minimal false predictions. These quantitative and qualitative results suggest that our approach provides adaptable, high-accuracy predictions in dynamic distributed systems and establishes a strong benchmark for link prediction in evolving network environments.

\begin{figure*}[ht]
    \centering
    \begin{subfigure}{0.19\textwidth}
        \centering
        \includegraphics[width=\textwidth]{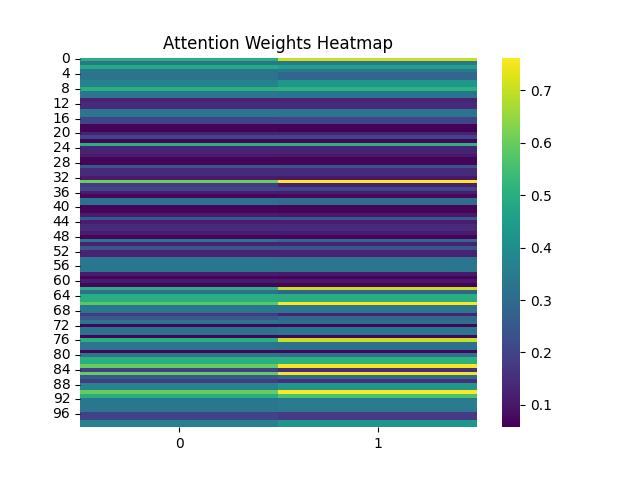}
        \caption{Epoch 0}
    \end{subfigure}
    \hfill
    \begin{subfigure}{0.19\textwidth}
        \centering
        \includegraphics[width=\textwidth]{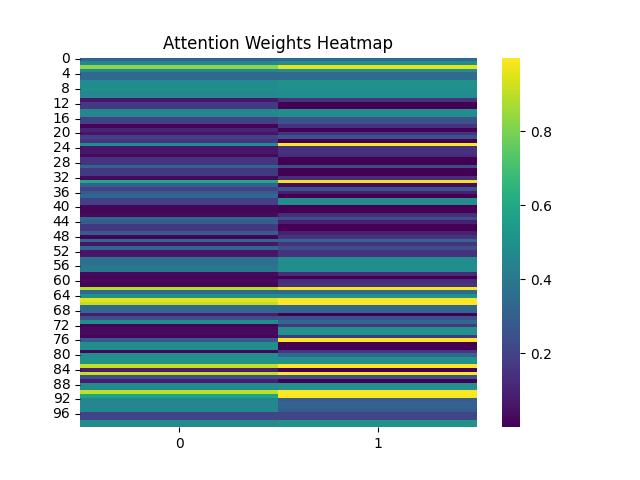}
        \caption{Epoch 49}
    \end{subfigure}
    \hfill
    \begin{subfigure}{0.19\textwidth}
        \centering
        \includegraphics[width=\textwidth]{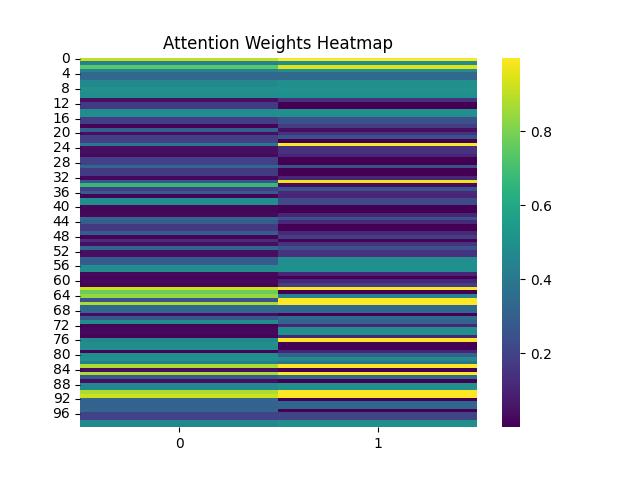}
        \caption{Epoch 99}
    \end{subfigure}
    \hfill
    \begin{subfigure}{0.19\textwidth}
        \centering
        \includegraphics[width=\textwidth]{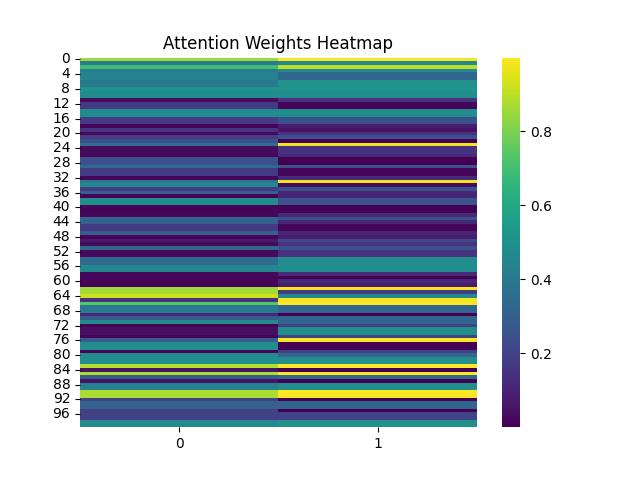}
        \caption{Epoch 149}
    \end{subfigure}
    \hfill
    \begin{subfigure}{0.19\textwidth}
        \centering
        \includegraphics[width=\textwidth]{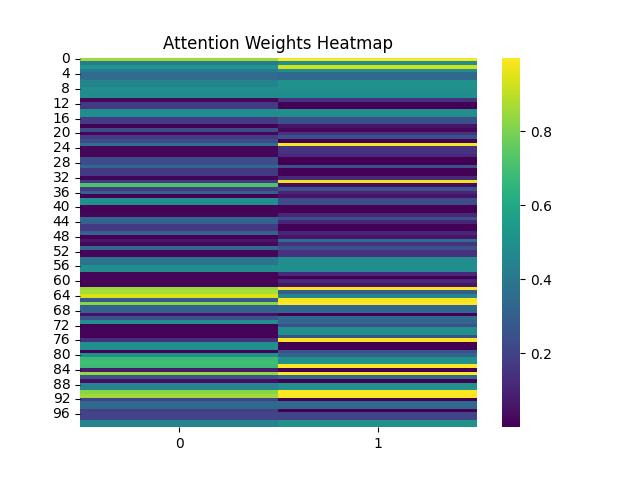}
        \caption{Epoch 199}
    \end{subfigure}
    \caption{Attention weight heatmaps for nodes 0-100 at selected training epochs (Epochs 0, 49, 99, 149, and 199). Each subfigure represents the attention distribution across nodes at the given epoch.}
    \label{fig:Nodes 0-100 heatmap}
\end{figure*}

\subsubsection{Qualitative Insights (Visual Analysis)}
To complement the quantitative evaluation, we gain additional insights through interpretative metrics like confusion matrixes, attention heatmaps, Precision-Recall (PR) curves, and Receiver Operating Characteristic (ROC) curves. These visualizations provide a detailed view of the model’s focus areas, error distribution, and predictive confidence over time, offering a nuanced understanding of its performance in dynamic microservice networks.

The confusion matrixes shown in Figure~\ref{fig:confusion_matrices} provide insights into the distribution of true positives, true negatives, false positives, and false negatives across different models. Our approach, which utilizes a GAT with temporal windowing and advanced negative sampling, demonstrates balanced sensitivity and specificity, minimizing false predictions more effectively than baseline methods like the Simple GNN, Simple Temporal GNN, and LSTM models.

In comparison, the LSTM model (Figure~\ref{fig:confusion_matrices}a) shows a high rate of both false positives and false negatives, indicating limitations in capturing the structural dependencies critical for microservice interactions. The Simple GNN (Figure~\ref{fig:confusion_matrices}b) improves on the structural insights but lacks temporal adaptability, resulting in imbalances between true positives and true negatives. The Simple Temporal GNN (Figure~\ref{fig:confusion_matrices}c) provides a better balance by incorporating temporal dynamics, yet it falls short in precision compared to our approach.

Our model (Figure~\ref{fig:confusion_matrices}d) achieves the best balance with a high number of true positives and true negatives, along with reduced false positives and false negatives. This indicates that our approach effectively distinguishes between actual and potential connections, highlighting its suitability for dynamic environments where prediction accuracy and stability are crucial.

In addition, we utilize attention heatmaps to visualize the evolution of attention weights assigned to specific edges over epochs, providing insights into how the model identifies critical relationships. Figure~\ref{fig:Nodes 0-100 heatmap} illustrates the model's attention progression, revealing marked shifts in focus during the initial training epochs, which become subtler as training progresses, indicating convergence. By Epoch 199, the attention weights stabilize, showing the model’s ability to capture consistent and significant interactions within the call graph. These heatmaps offer a deeper understanding of how the model prioritizes influential connections over time.

\begin{figure}[H]
    \centering
    \begin{subfigure}{0.45\columnwidth}
        \centering
        \includegraphics[width=\linewidth]{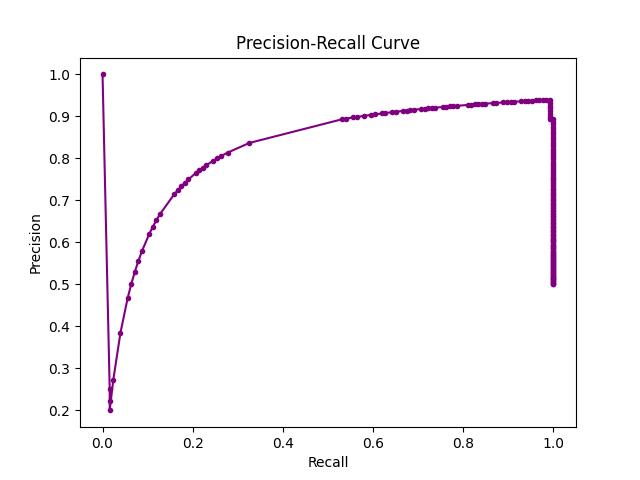}
        \caption{Precision-Recall Curve for Window 0.}
        \label{fig:PR_curve_window_0}
    \end{subfigure}
    \hfill
    \begin{subfigure}{0.45\columnwidth}
        \centering
        \includegraphics[width=\linewidth]{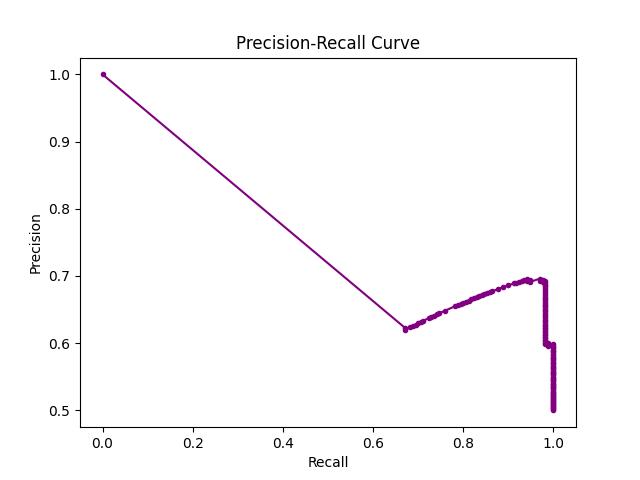}
        \caption{Precision-Recall Curve for Window 12.}
        \label{fig:PR_curve_window_12}
    \end{subfigure}
    \caption{Precision-Recall (PR) curves for representative time windows. The left panel illustrates Window 0, showcasing a smooth decline in precision as recall increases. The right panel depicts Window 12, with a sharper initial drop in precision, highlighting differences in model confidence across windows.}
    \label{fig:PR_Curve}
\end{figure}

The Precision-Recall (PR) curves, as depicted in Figure~\ref{fig:PR_Curve}, highlight the trade-off between precision and recall across varying confidence thresholds. For instance, Window 0 displays a smooth decline in precision as recall increases, indicating reliable model performance. In contrast, Window 12 exhibits a sharper initial drop in precision, which may suggest slight overconfidence in certain predictions during that interval. The overall consistency across multiple windows demonstrates the robustness of our model, even under varying conditions.

\begin{figure}[H]
    \centering
    \includegraphics[width=\columnwidth]{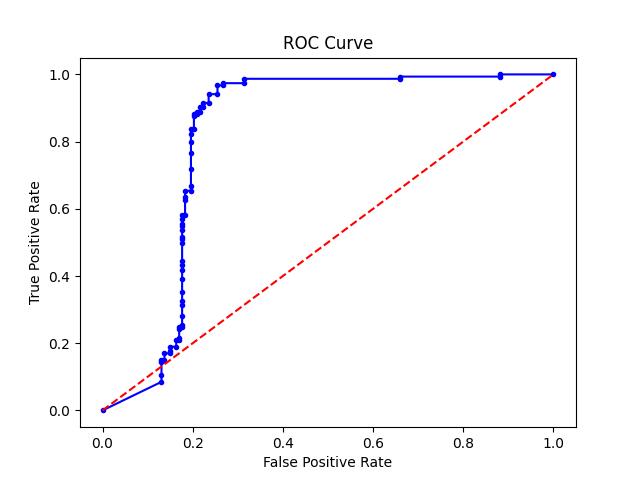}
    \caption{Receiver Operating Characteristic (ROC) curve for Time Window 21. The curve demonstrates a smooth progression from the bottom left to the top right, with an AUC approaching 1, indicating the model's strong capability to distinguish between positive and negative links. This pattern is consistent across multiple windows, highlighting the model’s reliable discrimination power for link prediction.}
    \label{fig:ROC_Curve}
\end{figure}

Lastly, we examine the Receiver Operating Characteristic (ROC) curve, an essential metric to evaluate model classification performance. The ROC curve in Figure~\ref{fig:ROC_Curve} displays a smooth progression from the bottom left to the top right, with an Area Under the Curve (AUC) approaching 1, indicating the model’s strong capability to distinguish between positive and negative links across thresholds. This pattern was consistent across multiple windows, further attesting to our model's reliable discrimination power for link prediction.

Overall, these visual insights complement our quantitative findings, illustrating that our model captures and emphasizes important relationships within the dynamic CallGraph dataset. Through confusion matrixes, attention heatmaps, PR curves, and ROC curves, we observe how the model distinguishes critical interactions, aligns with observed link patterns, and maintains high precision and recall across varied conditions. This qualitative analysis supports our approach as a well-rounded solution for link prediction tasks in dynamic, distributed environments.

\subsection{Discussion}

This section delves into the implications of our findings, highlighting the unique advantages and limitations of our approach in comparison to established methods. Our model's combination of temporal segmentation, advanced negative sampling, and attention mechanisms has proven effective for capturing the intricate dynamics of microservice networks, as demonstrated by both quantitative metrics and visual analyses. Here, we discuss the strengths of our approach, its interpretability, limitations, and potential for broader applicability.

\subsubsection{Comparative Performance and Key Insights}

Our model excelled in metrics such as F1 score (0.92), precision (0.89), and recall (0.96), showcasing its capability to capture complex, evolving interaction patterns. This performance contrasts sharply with structure-based methods like NodeSim, which achieved only an F1 score of 0.18, underscoring the limitations of static approaches in dynamic settings \cite{saxena2022nodesim}. While LSTM models effectively capture temporal patterns, their lack of structural context limits predictive accuracy in microservice networks. Similarly, Simple Temporal GNNs incorporate temporal insights but struggle with rapid network changes. Our model's temporal segmentation and advanced sampling address these gaps, resulting in a more nuanced adaptation to evolving network dynamics.

\subsubsection{Enhanced Interpretability Through Visual Analysis}

The model's interpretability is enhanced through attention heatmaps, PR curves, and ROC curves, which provide insights into its decision-making process. The attention heatmaps reveal the model's ability to prioritize critical connections over irrelevant ones, while confusion matrixes confirm a balanced classification of true positives and true negatives, crucial for high-stakes link prediction tasks. The consistency of PR and ROC curves across time windows further underscores the model's robustness, showing it maintains reliable precision and recall despite fluctuating network conditions. This interpretability strengthens our model’s potential as a tool for monitoring dynamic networks where understanding prediction rationale is vital.

\subsubsection{Challenges and Computational Constraints}

While effective, our model faced challenges in handling dataset imbalance and the computational demands of GNN training. Advanced negative sampling mitigated some issues related to imbalance, but extreme class disparities may still affect performance, suggesting potential for optimization in such cases. Moreover, the GNN’s computational demands remain higher than simpler alternatives, which could limit scalability. Future work could explore strategies for reducing computational overhead, such as using sparse representations or lighter GNN architectures, to improve efficiency without sacrificing accuracy.

\subsubsection{Generalizability and Broader Application Potential}

The design choices of our model—temporal segmentation, attention mechanisms, and advanced sampling—demonstrate strong adaptability across dynamic networks, potentially extending to other domains like social network analysis, fraud detection, and recommendation systems. Its ability to tune interaction strengths for varied datasets positions it as a versatile tool for diverse link prediction tasks. Future research can explore adapting this approach to other distributed systems, further validating its flexibility and broad applicability.

In summary, our model’s success underscores the importance of integrating temporal and structural insights to capture dynamic relationships effectively. By building on this foundation with further optimization and evaluation across broader datasets, our approach can serve as a robust solution for predictive tasks in complex distributed systems.

\subsection{Threats to Validity}
This study assesses the performance of our GNN-based approach for link prediction using a real-world dataset. Here, we discuss factors that may influence the generalizability and validity of our findings.

Firstly, we focused on AUC, Precision, Recall, and F1 Score to evaluate our model, as these metrics provide comprehensive insights into binary classification accuracy for identifying link existence. Unlike Mean Reciprocal Rank (MRR) or Hits@K, which assess ranking accuracy, these metrics align with our primary objective: accurately distinguishing between positive and negative links. Ranking metrics may be considered in future work, especially if prioritizing or ranking link predictions becomes a focus

Secondly, although our dataset is extensive, spanning a 10,000 ms interval, the results are specific to this chosen duration. An expanded dataset could provide deeper insights into the model’s ability to capture longer-term interactions and adapt to prolonged temporal dynamics, which are essential for applications requiring sustained predictive accuracy. Future studies could benefit from extending this temporal range to evaluate the model’s scalability and robustness over longer or more varied time periods.

Thirdly, although our evaluation is grounded in a large, comprehensive, and noisy real-world dataset, it is limited to a single dataset. This may restrict the generalizability of our findings to other datasets or application domains. Future work could address this by evaluating the model across multiple datasets, enhancing its potential applicability to a broader range of dynamic network contexts.

Lastly, the inherent computational demands of GNNs introduce a trade-off between model complexity and efficiency, especially when compared to simpler models. While our primary objective was to enhance predictive accuracy, future research could explore strategies for reducing computational costs, such as incorporating sparse representations or lightweight GNN architectures, without compromising performance.

\label{conclusions}
\section{Conclusions and Future Work}

Our study demonstrates the effectiveness of a GAT, a type of GNN, in accurately predicting link formation within microservice architectures. By capturing both structural and temporal dynamics, our model provides robust insights that enable early detection of potential dependencies, enhancing system resilience. Our comparative analysis underscores the advantage of GATs over other models, as reflected in our model’s high precision, recall, and balanced predictive performance across complex microservice networks. These findings establish GATs as a valuable tool for predictive tasks in distributed systems.

Future work could focus on enhancing the model’s scalability and efficiency, making it better suited for larger datasets and real-time applications. Incorporating additional metrics—such as interaction frequency, service load, or refined temporal patterns—could enrich the model's contextual understanding, potentially improving both predictive accuracy and interpretability. Architectural refinements and distributed training techniques may help manage computational demands, while validation across diverse datasets and metrics would further demonstrate its adaptability and robustness. Additionally, integrating advanced unsupervised or self-supervised learning techniques could improve generalization, especially for unseen data. With these advancements, our approach could become a powerful tool for accurate link prediction across various data-driven fields, from distributed systems to other complex networks.

Beyond this study, our approach holds significant potential for broader applications in predictive maintenance and adaptive monitoring within microservice ecosystems. By using link prediction to anticipate critical dependencies, adaptive monitoring systems can proactively address performance bottlenecks and potential service failures, enhancing overall system resilience. In this context, our model could serve as a foundation for proactive logging, tracing, and other diagnostic tools that dynamically adjust to evolving microservice interactions. These capabilities could extend to applications in network security and anomaly detection, where early detection and stability are essential.

\bibliographystyle{ACM-Reference-Format}
\bibliography{references}

\end{document}